\documentclass{article} % For LaTeX2e
\usepackage{iclr2021_conference,times}

\usepackage{amsmath,amsfonts,bm}

\def\eqref#1{equation~\ref{#1}}
\def\1{\bm{1}}

\DeclareMathAlphabet{\mathsfit}{\encodingdefault}{\sfdefault}{m}{sl}
\SetMathAlphabet{\mathsfit}{bold}{\encodingdefault}{\sfdefault}{bx}{n}

\usepackage{hyperref}
\usepackage{url}
\usepackage{graphicx}
\usepackage{subfigure}
\usepackage{CJKutf8}
\newcommand{\tabincell}[2]{\begin{tabular}{@{}#1@{}}#2\end{tabular}}

\title{MVP-BERT: Redesigning Vocabularies for chinese BERT and multi-vocab pretraining}

\author{Wei Zhu \\
Department of Computer Science\\
East China Normal University\\
Shanghai, China \\
\texttt{52205901018@stu.ecnu.edu.cn} \\
}

\iclrfinalcopy % Uncomment for camera-ready version, but NOT for submission.
\begin{document}

\maketitle

\begin{abstract}
Despite the development of pre-trained language models (PLMs) significantly raise the performances of various Chinese natural language processing (NLP) tasks, the vocabulary for these Chinese PLMs remain to be the one provided by Google Chinese Bert~\cite{devlin2018bert}, which is based on Chinese characters. Second, the masked language model pre-training is based on a single vocabulary, which limits its downstream task performances. In this work, we first propose a novel method, \emph{seg\_tok}, to form the vocabulary of Chinese BERT, with the help of Chinese word segmentation (CWS) and subword tokenization. Then we propose three versions of multi-vocabulary pretraining (MVP) to improve the models expressiveness.   Experiments show that: (a) compared with char based vocabulary, \emph{seg\_tok} does not only improves the performances of Chinese PLMs on sentence level tasks, it can also improve efficiency; (b) MVP improves PLMs' downstream performance, especially it can improve \emph{seg\_tok}'s performances on sequence labeling tasks.

\end{abstract}

\section{Introduction}

The pretrained language models (PLMs) including BERT~\cite{devlin2018bert} and its variants \cite{xlnet,roberta} have been proven beneficial for many natural language processing (NLP) tasks, such as text classification, question answering \cite{squad} and natural language inference (NLI)~\cite{bowman2015nli}, on English, Chinese and many other languages. Despite they brings amazing improvements for Chinese NLP tasks, most of the Chinese PLMs still use the vocabulary (vocab) provided by Google Chinese Bert~\cite{devlin2018bert}. Google Chinese Bert is a character (char) based model, since it splits the Chinese characters with blank spaces. In the pre-BERT era, a part of the literature on Chinese natural language processing (NLP) first do Chinese word segmentation (CWS) to divide text into sequences of words, and use a word based vocab in NLP models \cite{Word-Embedding-Composition,zou-etal-2013-bilingual}. There are a lot of arguments on which vocab a Chinese NLP model should adopt. 

The advantages of char based models are clear. First, Char based vocab is smaller, thus reducing the model size. Second, it does not rely on CWS, thus avoiding word segmentation error, which can directly result in performance gain in span based tasks such as named entity recognition (NER). Third, char-based models are less vulnerable to data sparsity or the presence of out-of-vocab (OOV) words, and thus less prone to over-fitting (\cite{is-word-seg-necessary}). However, word based model has its advantages. First, it will result in shorter sequences than char based counterparties, thus are faster. Second, words are less ambiguous, thus may be helpful for models to learn the semantic meanings of words. Third, with a word based model, exposure biases may be reduced in text generation tasks (\cite{zhao2013empirical}). Another branch of literature try to strike a balance between the two by combining word based embedding with character based embedding \cite{yin-etal-2016-multi,lstm-radical-level-features}.

\begin{figure}[h]
\begin{center}
%\framebox[4.0in]{$\;$}
\includegraphics[width=0.85\textwidth]{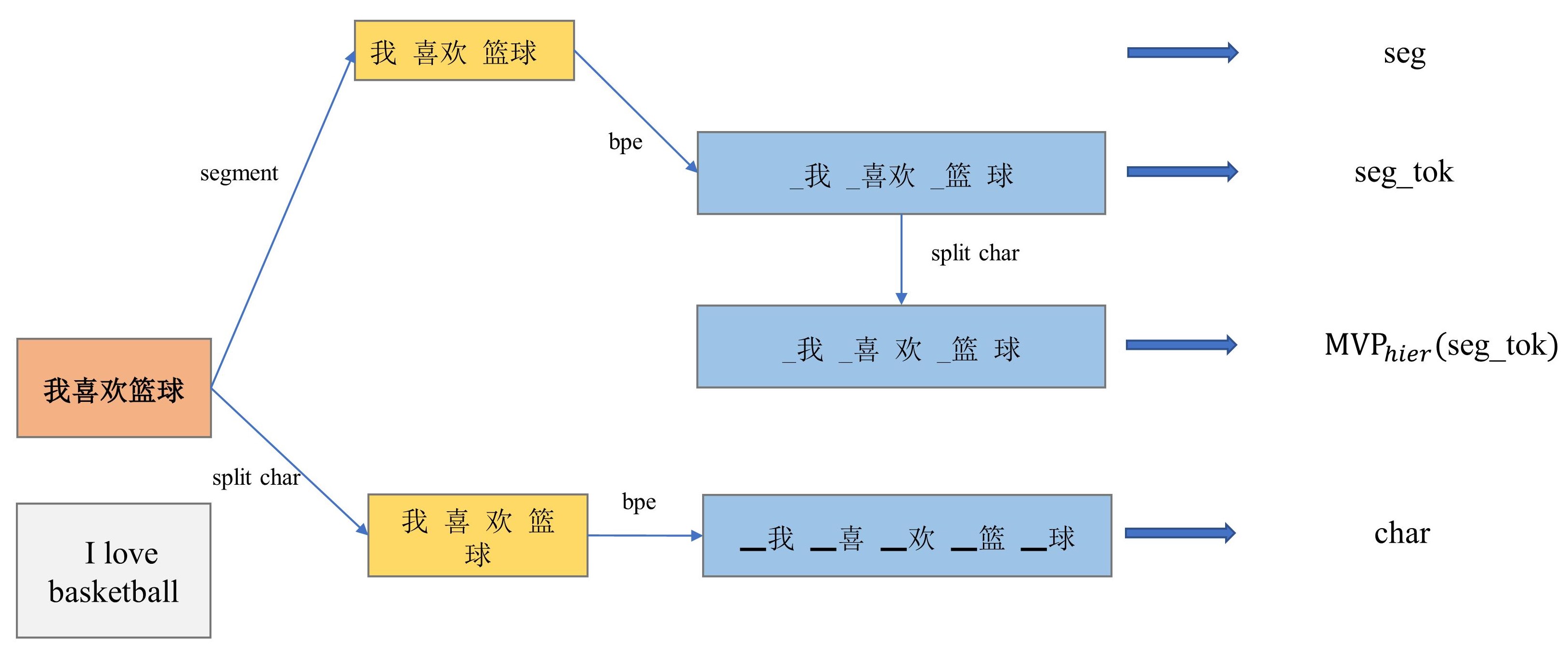}
\end{center}
\caption{An illustration of how to process input sentence into tokens under different methods we define.}
\label{fig:worflow}
\end{figure}

In this article, we try to strike a balance between the char based model and word based model and provides alternative approaches for building a vocab for Chinese PLMs. In this article, there are three approaches to build a vocab for Chinese PLMs: (1) following \cite{devlin2018bert}, separate the Chinese characters with white spaces, and then learn a sub-word tokenizer (denote as \emph{char}); (2) first segment the sentences with a CWS toolkit like jieba\footnote{https://github.com/fxsjy/jieba}, and then learn a sub-word tokenizer (denoted as \emph{seg\_tok}); (3) do CWS and kept the high-frequency words as tokens and low-frequency words will be tokenized by \emph{seg\_tok} (denoted as \emph{seg}). See Figure~\ref{fig:worflow} for their workflow of processing an input sentence. Note that the first one is essentially the same with the vocab of Google Chinese BERT. 

Inspired by the previous work that incorporate multiple vocabularies (vocabs) or combine multiple vocabs in an natural way\cite{yin-etal-2016-multi,lstm-radical-level-features}, we also investigate a series of strategies, which we will call Multi-Vocab Pretraining (MVP) strategies. The first version of MVP is to incorporate a hierarchical structure to combine the char based vocab and word based vocab. From the viewpoint of model forward pass, the embeddings of Chinese characters are aggregated to form the vector representations of multi-gram words or tokens, which then are fed into transformer encoders, and then the word based vocab will be used in masked language model (MLM) training. We will denote this version of MVP as MVP$_{hier}$. Note that in MVP$_{hier}$, the char based vocab is built by splitting the Chinese words in the word based vocab into Chinese chars, and non-Chinese tokens are kept the same. We will denote this strategy as MVP$_{hier}(V)$, where $V$ is a word based vocab. 

\begin{figure}[tb!]
\centering
\subfigure[MVP$_{hier}$]{%
\includegraphics[width=0.32\textwidth]{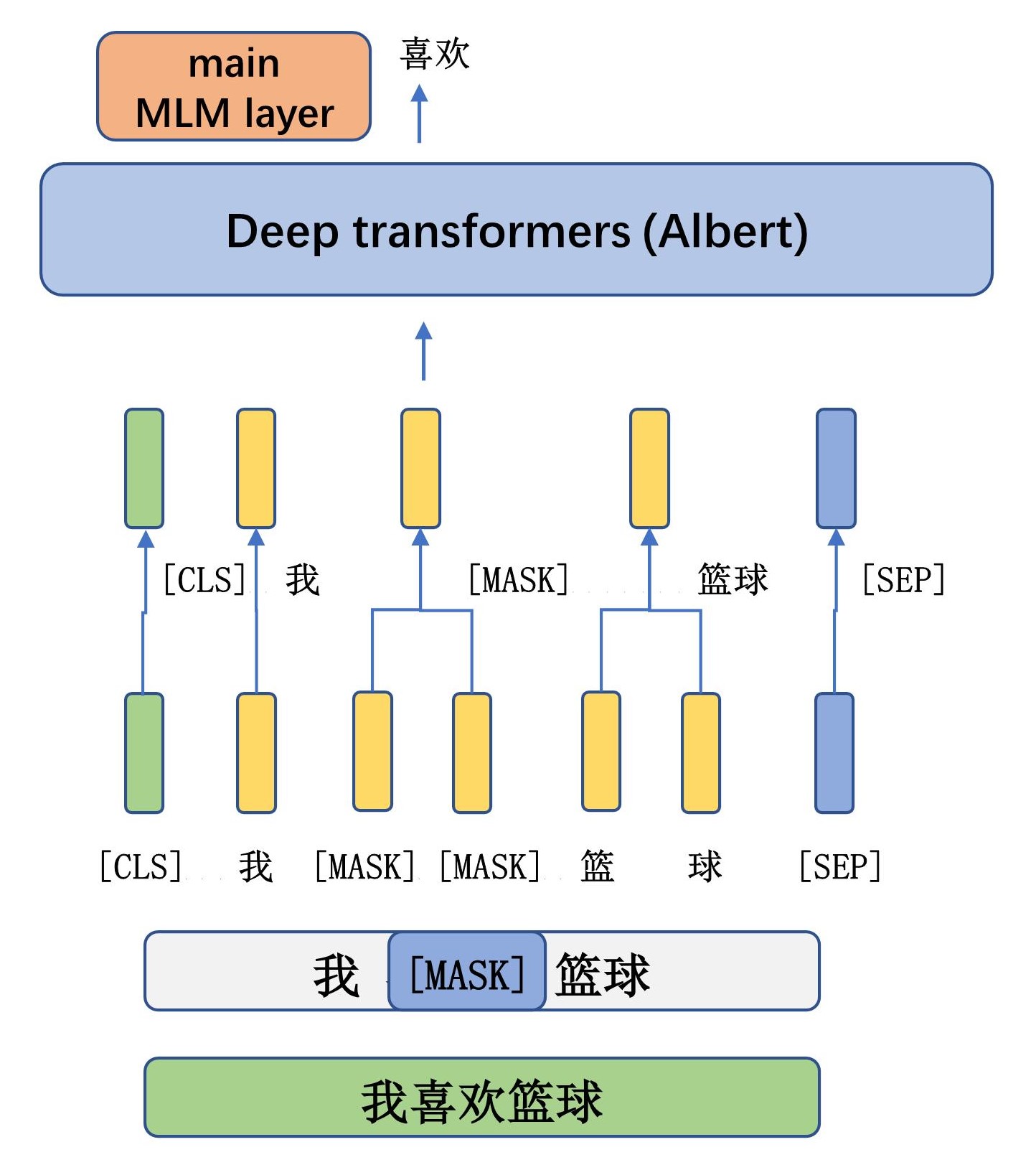}
\label{subfig:mvp_hier}}
\subfigure[MVP$_{obj}$]{%
\includegraphics[width=0.32\textwidth]{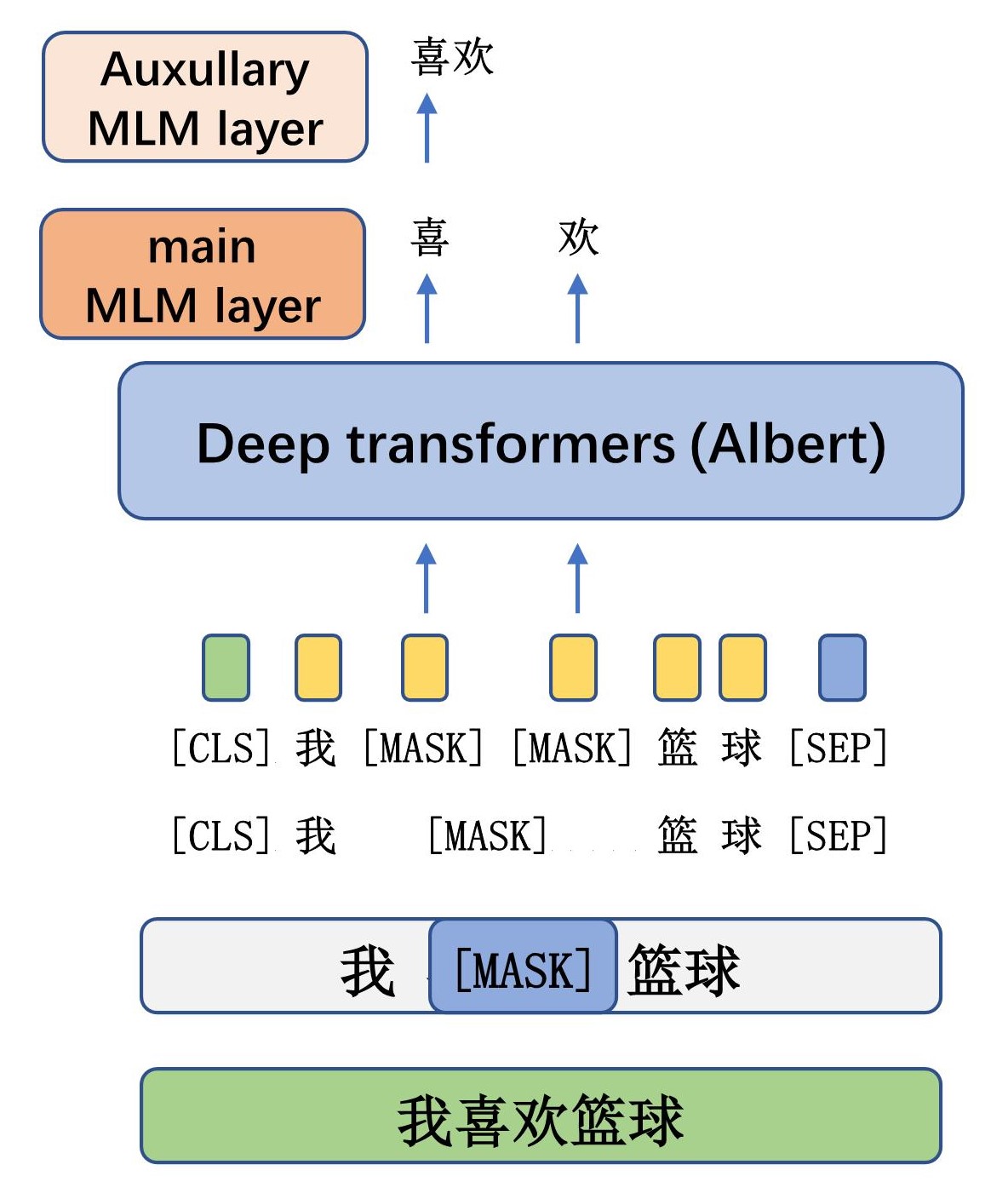}
\label{subfig:mvp_obj}}
\subfigure[MVP$_{pair}$]{%
\includegraphics[width=0.60\textwidth]{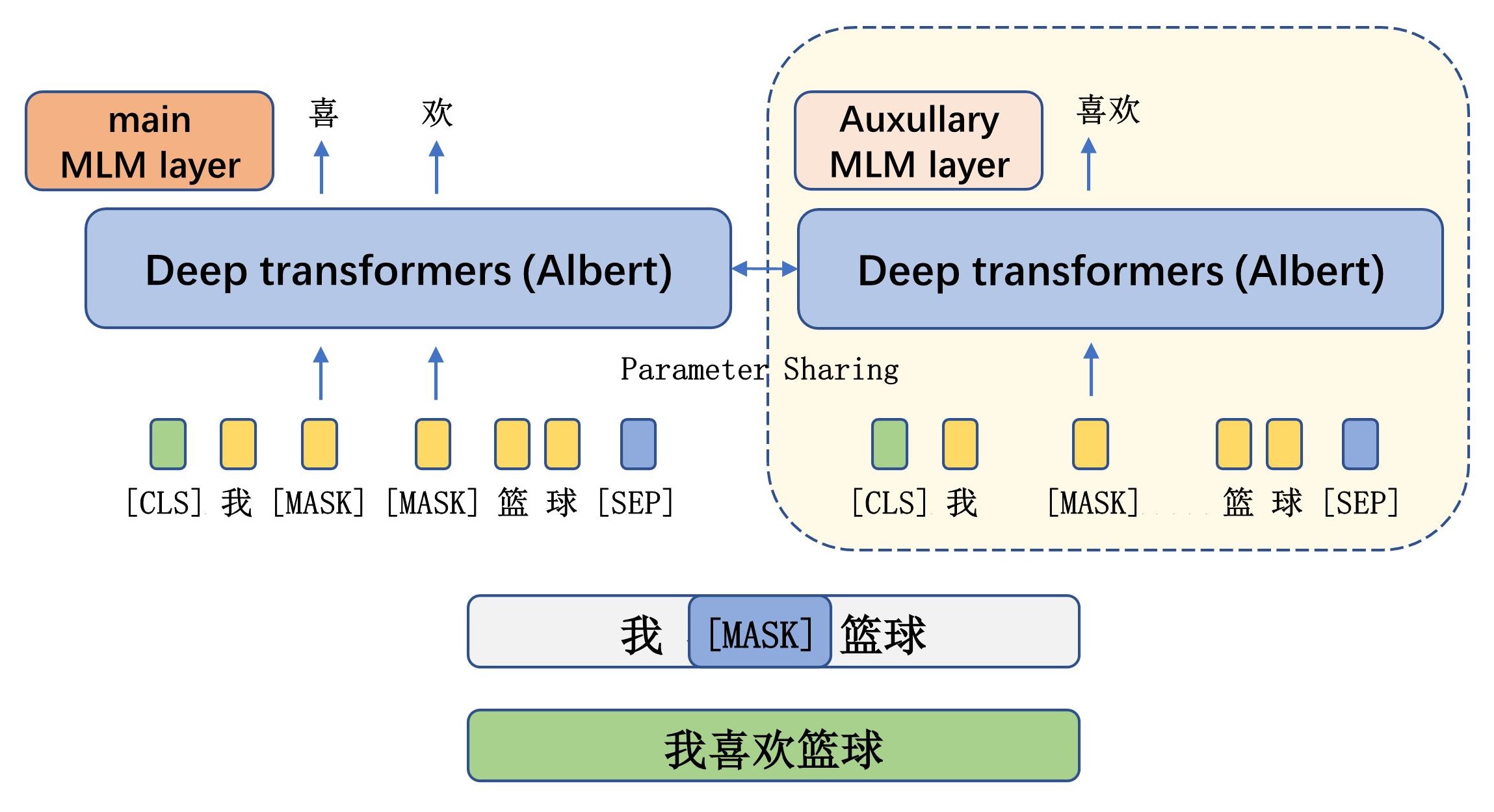}
\label{subfig:mvp_pair}}
\caption{The architectures for the three versions of MVP strategies.}
\label{fig:mvp}
\end{figure}

\begin{CJK*}{UTF8}{gbsn}
The second version of MVP (denoted as MVP$_{pair}$) is to employ a pair of vocabs in MLM. Due to limited resources, in this article we only consider the pair between \emph{seg\_tok} and \emph{char}. MVP$_{pair}$ is depicted in Figure \ref{subfig:mvp_pair}. In MVP$_{pair}$, a sentence (or a concatenation of multiple sentences in pre-training), is processed and tokenized both in \emph{seg\_tok} and \emph{char}, and the two sentences are encoded by two parameter-sharing transformer encoders. Whole word masking~\cite{wwm19chinese} is applied for pretraining. For example, the word "篮球" (basketball) is masked. The left encoder, which is with \emph{seg\_tok}, has to predict the single masked token is "篮球", and the right encoder has to predict "篮" and "球" for two masked tokens. MLM loss from both sides will be added with weights. With MVP$_{pair}$, parameter sharing enables the single vocab model to absorb information from the other vocab, thus enhancing its expressiveness. Note that after pre-training, one can either keep one of the encoder or both encoders for downstream finetuning. We will denote this strategy as MVP$_{pair}(V_1, V_2, i)$, where $V_1$ and $V_2$ are two different vocabs, $i=s$ means only the encoder with $V_1$ is kept for finetuning (single vocab model), and $i=e$ means both encoders are kept (ensemble model).  

The third version of MVP (denoted as MVP$_{obj}$) is depicted in Figure \ref{subfig:mvp_obj}. In MVP$_{obj}$, the sentence is encoded only once with a fine-grained vocab, and MLM task with that vocab is conducted. As in the figure, he word "喜欢" (like) is masked, and under the vocab of \emph{char}, the PLM  has to predict "喜" and "欢" for the two masked tokens. As additional training objective, we will employ a more coarse-grained vocab like \emph{seg\_tok} and ask the model to use the starting token ("喜")'s representation to predict the original word under \emph{seg\_tok}. We will denote this strategy as MVP$_{pair}(V_1, V_2)$, where $V_1$ and $V_2$ are a pair of vocabs and $V_1$ is the more fine-grained one. 

\end{CJK*}

Extensive experiments and ablation studies are conducted. We select BPE implemented by sentencepiece\footnote{https://github.com/google/sentencepiece} as the sub-word model, and Albert~\cite{lan2019albert} (base model) as our PLM. Pre-training is done on Chinese Wikipedia corpus\footnote{https://dumps.wikimedia.org/zhwiki/latest/}, which is also the corpus on which we build the different vocabs. After pretraining, the three vocab building methods and three MVP strategies are compared on a series of Chinese benchmark datasets, 4 of which are sentence classification tasks, 3 are sequence labeling tasks. The experimental results demonstrate the following take-aways: 1) directly use Chinese words as vocab (\emph{seg}) does not perform well; 2) \emph{seg\_tok} based ALBERT consistently performs better than \emph{char} and \emph{seg} on sentence classification tasks, but not sequence labeling tasks; 3) MVP strategies can help to improve a single vocab model on both types of tasks, especially it can help \emph{seg\_tok} on sequence labeling tasks; 4) MVP$_{pair}$ ensemble are the best model, but it comes with a higher inference time; 5) MVP$_{obj}$ help to provide a better single vocab model than MVP$_{pair}$.

We now summaries the following contributions in this work. 

\begin{itemize}
    \item experiments on three ways of building new vocab for Chinese BERT, and \emph{seg\_tok}, the combination of CWS and subword tokenization is novel. 
    \item We propose 3 MVP strategies for enhancing the Chinese PLMs.  
\end{itemize}

\section{RELATED WORK}

Before and since~\cite{devlin2018bert}, a large amount of literature on pretrained language model appear and push the NLP community forward with a speed that has never been witnessed before. \citet{Peters2018elmo} is one of the earliest PLMs that learns contextualized representations of words. GPTs~\cite{radford2018language,radford2019language} and BERT~\cite{devlin2018bert} take advantages of Transformer~\cite{vaswani2017attention}. GPTs are uni-directional and make prediction on the input text in an auto-regressive manner, and BERT is bi-directional and make prediction on the whole or part of the input text. In its core, what makes BERT so powerful are the pretraing tasks, i.e., Mask language modeling (MLM) and next sentence prediction (NSP), where the former is more important. Since BERT, a series of improvements have been proposed. The first branch of literature improves the model architecture of BERT. ALBERT~\cite{lan2019albert} makes BERT more light-weighted by embedding factorization and progressive cross layer parameter sharing. \citet{bigbird} improve BERT's performance on longer sequences by employing sparser attention. 

The second branch of literature improve the training of BERT. \cite{roberta} stabilize and improve the training of BERT with larger corpus. More work have focused on new language pretraining tasks. ALBERT~\cite{lan2019albert} introduce sentence order prediction (SOP). StructBERT~\cite{StructBERT} designs two novel pre-training tasks, word structural task and sentence structural task, for learning of better representations of tokens and sentences. ERNIE 2.0~\cite{ERNIE2} proposes a series of pretraining tasks and applies continual learning to incorporate these tasks. ELECTRA~\cite{electra} has a GAN-style pretraining task for efﬁciently utilizing all tokens in pre-training. Our work is closely related to this branch of literature by design a series of novel pretraining objective by incorporating multiple vocabularies. Our proposed tasks focus on intra-sentence contextual learning, and it can be easily incorporated with other sentence structural tasks like SOP. 

Another branch of literature look into the role of words in pre-training. Although not mentioned in \cite{devlin2018bert}, the authors propose whole word masking in their open-source repository, which is effective for pretraining BERT. In SpanBERT~\cite{joshi2019spanbert}, text spans are masked in pre-training and the learned model can substantially enhance the performances of span selection tasks. It is indicated that word segmentation is especially important for Chinese PLMs. \citet{chinese-bert-wwm} and \citet{sun2019ernie} both show that masking tokens in the units of natural Chinese words instead of single Chinese characters can significantly improve Chinese PLMs. In this work, compared to literature, we propose to re-design the vocabulary of the Chinese BERT by combining word segmentation and sub-word tokenizations.

\section{Our methods}

In this section, we present our methods for rebuilding the vocab for Chinese PLMs, and introduce our series of MVP strategies. 

\subsection{Building the vocabs}

We investigate four work-flows to process the text inputs, each corresponding to a different vocab (or a group of vocabs) (Figure \ref{fig:worflow}). We first introduce the single vocab models, \emph{char}, \emph{seg\_tok} and \emph{seg}. For char based vocab \emph{char}, Chinese characters in the corpus are treated as words in English and are separated with blank spaces and a sub-word tokenizer is learned. \emph{seg\_tok} requires the sentences in corpus to be segmented and a sub-word tokenizer like BPE are learned on the segmented sentences. Note that in \emph{seg\_tok}, some natural word will be splitted into pieces, but there are still many tokens that have multiple Chinese chars. Finally, \emph{seg} with size $N$ is built with the following procedures: a) do CWS on the corpus; b) count the words' frequency, and add the tokens from \emph{seg\_tok} with frequency 0; c) for long tail Chinese words and non-Chinese tokens, tokenize them into subword tokens with \emph{seg\_tok}, and the words' frequencies are added to the sub-word tokens' frequencies; d) sort the vocab via frequency, and if the most frequent N words or tokens can cover 99.95\% of the corpus, then take them as vocab. Note that some of the tokens from \emph{seg\_tok} will be dropped. 

\subsection{Multi-vocab pretraining (MVP)}

In this subsection, we will introduce MVP, which is a derivation of MLM task by~\citet{devlin2018bert}. MVP has three versions, MVP$_{hier}$, MVP$_{obj}$ and MVP$_{pair}$, all of which have one thing in common, that is, they require more than one vocab to implement pre-training. 

\begin{CJK*}{UTF8}{gbsn}
Figure \ref{subfig:mvp_hier} depicts the architecture of MVP$_{hier}$ and Figure~\ref{subfig:mvp_hier} depicts its procedure for processing input sentences. Two vocab, a fine-grained vocab $V_f$, and a more coarse-grained vocab $V_c$, are combined in a hierarchical way. Sequences are first tokenized via $V_c$, and then the Chinese tokens (if containing multiple Chinese characters) are splitted into single characters. Chinese characters and non-Chinese tokens are embedded into vectors. Then representations of chars inside a word is aggregated into the representation of this word, which is further fed into the transformer encoder. During MLM task, whole word masking is applied, that is, we will mask 15\% of the tokens in the $V_c$. For example in Figure \ref{subfig:mvp_hier}, "\_喜欢" (like) is masked, thus in the char sequence, two tokens "\_喜" and "欢" are masked. Then an aggregator will combine the embeddings into the vectors of word tokens. At the MLM task, a classifier is designated to predict the masked word token "\_喜欢", which is from $V_c$. Let $\mathbf{x}$ and $\mathbf{y}$ denote the sequences of tokens with length $l_x$ and $l_y$, for the same sentence under $V_c$ and $V_f$, in which a part of tokens are masked. Denote $\mathbf{x}^{mask}$ as the masked tokens under $V_c$. The loss function for MVP$_{hier}$ is  
\begin{equation*}
    \min_{\theta} - \log \mathrm{P}_{\theta}(\mathbf{x}^{mask} | \mathbf{x}, \mathbf{y}) \approx  \min_{\theta} - \sum_{i=1}^{l_x} \mathrm{I}_{i}\log \mathrm{P}_{\theta}(x_{i}^{mask} | \mathbf{x}, \mathbf{y}),
\end{equation*}
in which $\mathrm{I}_{i}^{x}$ is a variable with binary values indicating whether the $i$-th token is masked in $\mathbf{x}$.  

\end{CJK*}

\begin{CJK*}{UTF8}{gbsn}
In MVP$_{obj}$, a sentence is tokenized and embedded in a fine-grained $V_f$(e.g., a char based vocab), and MLM task on $V_f$ is conducted. However, different from the vanilla MLM, another MLM task based on a more coarse-grained vocab $V_c$ is added. For example, encoded representations of the chars "\_喜" and "欢" inside the word "\_喜欢" is aggregated to the vector representation of the word, and an auxiliary MLM layer is tasked to predict the word based on $V_c$. For the aggregator in the example, we adopt BERT-style pooler, which is to use the starting token's representation to represent the word's representation.\footnote{Due to limited resources available, we leave to future work to investigate whether alternative aggregators can bring improvements. } Denote $\mathbf{x}^{mask}$ and $\mathbf{y}^{mask}$ as the masked tokens under $V_f$ and $V_c$ respectively. The loss function for MVP$_{obj}$ is as follows: 
\begin{equation*}
    \min_{\theta} - \log \mathrm{P}_{\theta}(\mathbf{x}^{mask}, \mathbf{y}^{mask} | \mathbf{x} ) \approx  \min_{\theta} - \sum_{i=1}^{l_x} \mathrm{I}^{x}_{i}\log \mathrm{P}_{\theta}(x_{i}^{mask} | \mathbf{x}) - \lambda * \sum_{i=1}^{l_y} \mathrm{I}^{y}_{i}\log \mathrm{P}_{\theta}(y_{i}^{mask} | \mathbf{x}),
\end{equation*}
in which $\mathrm{I}^{x}_{i}$ and $\mathrm{I}^{y}_{i}$ are variables with binary values indicating whether the $i$-th token is masked in sequence $\mathrm{x}$ and $\mathrm{y}$ respectively. Here $\lambda$ is the coefficient which measures the relative importance of the auxiliary MLM task. 

Now we introduce MVP$_{pair}$, which is the most resource-demanding version of MVP, but it will be proven beneficial. Our goal is to enhance the model with a single vocab, by introducing additional encoder with another vocab and a corresponding MLM task. In this strategy, a sentence is tokenized and embedded with two vocabs, $V_1$ and $V_2$, which will be fed into separate transformer encoders. Transformer encoders will share the same parameters, but the embedding layers are separate. For example in Figure \ref{subfig:mvp_pair}, "\_喜欢" is masked, and thus two tokens on the left sequence and(or) one token on the right sequence are masked. The left encoder and MLM layer is tasked to recover the two tokens "\_喜" and "欢". And on the right, the other MLM layer needs to recover "\_喜欢". Through parameter sharing, self-supervised signals from one vocab is transferred to the model with the other vocab. Formally, Thus the loss function for  MVP$_{pair}$ is
\begin{equation*}
    \min_{\theta} - \log \mathrm{P}_{\theta}(\mathbf{x}^{mask}, \mathbf{y}^{mask} | \mathbf{x}, \mathbf{y} ) \approx  \min_{\theta} - \sum_{i=1}^{l_x} \mathrm{I}^{x}_{i}\log \mathrm{P}_{\theta}(x_{i}^{mask} | \mathbf{x}) - \lambda * \sum_{i=1}^{l_y} \mathrm{I}^{y}_{i}\log \mathrm{P}_{\theta}(y_{i}^{mask} | \mathbf{y}),
\end{equation*}

If after MVP$_{pair}$ pretraining we decide to only keep one of the encoder, we will call this model single vocab MVP$_{pair}$. Otherwise, we can call the model as ensemble MVP$_{pair}$. For single mode MVP$_{pair}$, finetuning is the same with vanilla ALBERT. Finetuning for ensemble mode MVP$_{pair}$ is different. For sentence classification, the pooled vectors on both [CLS] tokens will be concatenated to be the feature vector of the classifier. When doing sequence labeling tasks, two ways of ensemble can be conducted, which we will use the example from Figure \ref{subfig:mvp_pair} to illustrate. The first approach is to concatenate the features from fine-grained encoder to the coarse-grained encoder. That is, representations of "\_喜" and "欢" are aggregated to the representation of "\_喜欢", and it will be concatenated to the representation from the coarse-grained encoder on the right. We will call this approach fine-to-coarse ensemble. The other approach is coarse-to-fine ensemble, which is to concatenate the representation of "\_喜欢" from the coarse-grained encoder to "\_喜" and "欢" from the fine-grained encoder. Then the labels of "\_喜" and "欢" are predicted.

\end{CJK*}

For notational convenience, we will denote the model pretrained with MVP$_{hier}$ strategy and vocab $V$ as MVP$_{hier}(V)$. MVP$_{obj}$ with a fine-grained vocab $V_{f}$ and a coarse-grained vocab $V_{c}$ are denoted as MVP$_{obj}(V_{f}, V_{c})$. MVP$_{pair}$ with two vocab $V_{1}$ and $V_{2}$ is denoted as MVP$_{pair}(V_{1}, V_{2}, i, j)$, where $i$ can be $single$, meaning only keep the encoder from $V_{1}$ for finetuning, or $ensemble$, meaning keep both encoders, and $j$ can be $ftc$ (short for fine-to-coarse) or $ctf$ (coarse-to-fine), which is two approaches for finetuning MVP$_{pair}$ on sequence labeling tasks. Note that if $i$ equals $single$, we will neglect the parameter $j$.

\section{Experiments}

\subsection{Setup}

For pre-training corpus, we use Chinese Wikipedia. The vocab size is 21128 for \emph{char} and \emph{seg\_tok}. 50.38\% of \emph{seg\_tok} are Chinese tokens with length more than 1. 

In this article, we adopt jieba as our CWS tool.\footnote{Despite there are more sophisticated CWS toolkits avalable, and using them may lead to better performances, jieba is efficient and good enough to prove the importance of word segmentation in the vocab design of Chinese BERT.} To keep 70\% of the vocab as natural words from CWS, and cover 99.95\% of the corpus, \emph{seg} is built with vocab size 69341. 

For MVP$_{hier}$, we consider MVP$_{hier}(seg\_tok)$. For MVP$_{obj}$, we consider MVP$_{obj}(char, seg\_tok)$ and MVP$_{obj}(seg\_tok, seg)$, with $\lambda$ equal to 0.1, 0.5, 1.0, 2.0, 10.0. For MVP$_{pair}$, we consider MVP$_{pair}(char, seg\_tok)$, the combination of \emph{seg\_tok} and char based vocab, with $\lambda$ equal to 0.1, 0.5, 1.0, 2.0, 10.0. Note that to maintain the consistency for non-Chinese tokens, the char based vocab in MVP$_{hier}(seg\_tok)$, MVP$_{obj}(char, seg\_tok)$ and MVP$_{pair}(char, seg\_tok)$  is derived from \emph{seg\_tok} by splitting Chinese characters of \emph{seg\_tok} tokens into characters, not exactly the same with \emph{char}. 

For pretraining, whole word masking is adopted, and total 15\% of the words (from CWS) in the corpus are masked, which are then tokenized into different tokens under different vocabs. For MVP$_{obj}(char, seg\_tok)$ and MVP$_{pair}(char, seg\_tok)$, 1/3 of the time only tokens \emph{seg\_tok} are predicted, and 1/3 of the time only tokens from the derived char based vocab are predicted, and for the rest of the time, tokens from both vocabs are predicted.

In this article, all models use the ALBERT as encoder. We make use of a smaller parameter settings, that is, the number of layer is 3, the embedding size is 128 and the hidden size is 256. Other ALBERT configurations remain the same with ALBERT \cite{lan2019albert}. The pretraining hyper-parameters are almost the same with ALBERT \cite{lan2019albert}. The maximum sequence length is 512. Here, the sequence length is counted under the coarse-grained vocab for MVP$_{hier}$,  fine-grained vocab for MVP$_{obj}$, and the longer one of the two sequences under the two vocabs for MVP$_{pair}$. The batch size is 1024, and all the model are trained for 12.5k steps. The pretraining optimizer is LAMB and the learning rate is 1e-4. For finetuning, the sequence length is 256, the learning rate is 2e-5, the optimizer is Adam~\cite{KingmaICML15} and the batch size is set as the power of 2 and so that each epoch contains less than 1000 steps. Each model is run on a given task for 10 times and the average performance scores and standard deviations are reported for reproducibility.

\subsection{benchmark tasks}

For downstream tasks, we select 4 text classification tasks: (1) ChnSentiCorp (chn)\footnote{https://github.com/pengming617/bert classification}, a hotel review dataset; (2) Book review\footnote{https://embedding.github.io/evaluation/} (book\_review), collected from Douban\footnote{https://book.douban.com/} by \cite{K-BERT}. For sentence pair classification tasks, we include the following 4 datasets: (1) XNLI (xnli) from~\cite{conneau-etal-2018-xnli}; (2) LCQMC (lcqmc)~\cite{liu-etal-2018-lcqmc}; (3) NLPCC-DBQA\footnote{http://tcci.ccf.org.cn/conference/2016/dldoc/evagline2.pdf} (nlpcc\_dbqa), a question-answer matching task in the open domain; (4) Law QA~\cite{K-BERT}, a QA matching task in the legal domain. We also investigate three NER tasks. MSRA NER (msra)~\cite{levow-2006-third} is from open domain, Finance NER\footnote{https://embedding.github.io/evaluation/\#extrinsic} (fin) is from financial domain, and CCKS NER\footnote{https://biendata.com/competition/CCKS2017\_2/} (ccks) is collected from medical records.

\subsection{Experimental results}

We first compare the three single vocab models. We find that word based vocabs \emph{seg\_tok} and \emph{seg} perform better than \emph{char} on all sentence classification tasks, and \emph{seg\_tok} outperforms \emph{seg} even though having less parameters. However, \emph{seg\_tok} is worse than or comparable to \emph{char} on three NER tasks, where these performance gaps are only partially explained by the CWS errors\footnote{CWS errors accounts for only less than 0.3\% of the errors on CCKS.}. We believe that its disadvantages on sequence labeling tasks is due to sparcity of tokens, i.e., each token has less training samples, thus are less well trained for token level classification. From the above results, we show that both CWS and subword tokenization are important to improve the expressiveness of Chinese PLMs in sentence level tasks via vocab re-designing. In addition, \emph{seg\_tok} is more efficient than \emph{seg\_tok}. we can observe a 1.35x speed up when changing the vocab from \emph{char} to \emph{seg\_tok}.

\begin{table*}
\centering
\resizebox{0.97\textwidth}{!}{
\begin{tabular}{cccccccccc}
\hline 
task & chn & br & lcqmc & xnli & nlpcc & msra & fin  & ccks   \\ 
\hline
metric  &    acc &  acc  & acc  &  acc  &  macro F1 &  exact F1  &  exact F1 &  exact F1 \\
 \hline
 
char  &   \tabincell{c}{86.61 \\ $\pm$ 1.12} & \tabincell{c}{77.83 \\ $\pm$ 0.55} & \tabincell{c}{77.85 \\ $\pm$ 0.78}  & \tabincell{c}{59.22 \\ $\pm$ 0.76}  &  \tabincell{c}{ 62.09 \\ $\pm$ 2.57}  & \tabincell{c}{\textbf{81.14} \\ $\pm$ 0.61} & \tabincell{c}{\textbf{73.43} \\ $\pm$ 0.61} &  \tabincell{c}{\textbf{85.63} \\ $\pm$ 0.28} \\

seg\_tok    &  \tabincell{c}{\textbf{87.17} \\ $\pm$ 0.46}           & \tabincell{c}{\textbf{79.08} \\ $\pm$ 0.20} & \tabincell{c}{\textbf{79.79} \\ $\pm$ 0.42} & \tabincell{c}{\textbf{60.19} \\ $\pm$ 0.43} & \tabincell{c}{\textbf{64.02} \\ $\pm$ 1.15} & \tabincell{c}{79.81 \\ $\pm$ 0.81} & \tabincell{c}{72.12 \\ $\pm$ 1.26}  & \tabincell{c}{84.79 \\ $\pm$ 0.74}   \\ 

seg  &   \tabincell{c}{87.06 \\ $\pm$ 0.78}  & \tabincell{c}{78.53 \\ $\pm$ 0.43} & \tabincell{c}{79.27 \\ $\pm$ 0.68} & \tabincell{c}{59.71 \\ $\pm$ 0.59} & \tabincell{c}{63.32 \\ $\pm$ 1.83} & \tabincell{c}{79.07 \\ $\pm$ 1.17} & \tabincell{c}{71.24 \\ $\pm$ 1.64}  & \tabincell{c}{83.96 \\ $\pm$ 0.93}   \\ 

\hline

MVP$_{hier}(seg\_tok)$   &   \tabincell{c}{87.06 \\ $\pm$ 0.48}  &  \tabincell{c}{78.67 \\ $\pm$ 0.53} &  \tabincell{c}{79.64 \\ $\pm$ 0.59}   &  \tabincell{c}{59.89 \\ $\pm$ 0.53}  &  \tabincell{c}{64.06 \\ $\pm$  1.32} &  \tabincell{c}{80.57 \\ $\pm$ 0. 75}  &  \tabincell{c}{72.36 \\ $\pm$ 1.07 }  &  \tabincell{c}{84.88 \\ $\pm$ 0.98}  \\

MVP$_{obj}(char, seg\_tok)$ &    
\tabincell{c}{87.26 \\ $\pm$ 0.52}  &  
\tabincell{c}{78.45 \\ $\pm$ 0.64}  &  
\tabincell{c}{78.92 \\ $\pm$ 0.56}  &  
\tabincell{c}{59.98 \\ $\pm$ 0.48} &  
\tabincell{c}{63.67 \\ $\pm$ 1.22}  &  
\tabincell{c}{81.56 \\ $\pm$ 0.56}  &  
\tabincell{c}{73.95 \\ $\pm$ 0.58}  &  
\tabincell{c}{86.10 \\ $\pm$ 0.35}  \\

MVP$_{obj}(seg\_tok, seg)$  &          \tabincell{c}{87.55 \\ $\pm$ 0.32}  &          \tabincell{c}{79.67 \\ $\pm$ 0.21} &          \tabincell{c}{80.44 \\ $\pm$ 0.39}  &        \tabincell{c}{60.57 \\ $\pm$ 0.36}  &        \tabincell{c}{65.48 \\ $\pm$ 1.24}  &         \tabincell{c}{81.05. \\ $\pm$ 0.77}  &        \tabincell{c}{73.47 \\ $\pm$ 1.23}  &        \tabincell{c}{85.56 \\ $\pm$ 0.68} \\

MVP$_{pair}(char, seg\_tok, s)$  &    \tabincell{c}{87.04 \\ $\pm$ 0.66}  &     \tabincell{c}{78.56 \\ $\pm$ 0.68}  &        \tabincell{c}{78.63 \\ $\pm$ 0.63}  &        \tabincell{c}{60.23 \\ $\pm$ 0.47}  &     \tabincell{c}{63.78 \\ $\pm$ 1.28}  &     \tabincell{c}{81.47 \\ $\pm$ 0.43}  &   \tabincell{c}{73.78 \\ $\pm$ 0.78}  &         \tabincell{c}{85.97 \\ $\pm$ 0.45} \\

MVP$_{pair}(seg\_tok, char, s)$  &    \tabincell{c}{87.22 \\ $\pm$ 0.41}  &          \tabincell{c}{79.43 \\ $\pm$ 0.31}  &          \tabincell{c}{80.05 \\ $\pm$ 0.44}  &         \tabincell{c}{60.25 \\ $\pm$ 0.52}  &         \tabincell{c}{64.79 \\ $\pm$ 1.03}  &     \tabincell{c}{80.03 \\ $\pm$ 0.85}  &         \tabincell{c}{73.17 \\ $\pm$ 1.18}  &     \tabincell{c}{85.45 \\ $\pm$ 0.47} \\

MVP$_{pair}(seg\_tok, char, e, ftc)$  &   \tabincell{c}{\textbf{87.96} \\ $\pm$ 0.37} & \tabincell{c}{\textbf{80.05} \\ $\pm$ 0.18} & \tabincell{c}{\textbf{80.56} \\ $\pm$ 0.32} & \tabincell{c}{\textbf{60.86} \\ $\pm$ 0.29} & \tabincell{c}{\textbf{65.94} \\ $\pm$ 0.94} & \tabincell{c}{81.89 \\ $\pm$ 0.65}  &     \tabincell{c}{73.97 \\ $\pm$ 0.56}  &         \tabincell{c}{86.27 \\ $\pm$ 0.36} \\   

MVP$_{pair}(char, seg\_tok, e, ctf)$  &           -  &                                         -  &                                         -  &                                         -  &                                         -  &                                  \tabincell{c}{\textbf{82.28} \\ $\pm$ 0.54}  & \tabincell{c}{\textbf{74.32} \\ $\pm$ 0.47}  & \tabincell{c}{\textbf{87.85} \\ $\pm$ 0.32} \\ 

\hline
\end{tabular}}
\caption{\label{tab:main_results} Main results on the Chinese benchmark datasets. For each task and each model, experiments are repeated for 10 times, and the average and standard deviation of the scores are reported.}
\end{table*}

The lower rows of Table \ref{tab:main_results} also demonstrates the effectiveness of MVP training. Note that in this table, we report MVP$_{obj}$ with $\lambda=2.0$, and MVP$_{pair}$ with $\lambda=1.0$. First, we can see MVPs consistently improve the models' performances, among which the largest improvements come from MVP$_{obj}$ and ensemble MVP$_{pair}$. Second, for \emph{seg\_tok}, keeping only one encoder provides improvements on sequence labeling tasks, but the improvements are less than that provided by MVP$_{pair}(seg\_tok, seg)$ on sentence level tasks. Third, note that for NER tasks, ensemble MVP$_{pair}$ performs better than single MVP$_{pair}$, and coarse-to-fine ensemble is more effective than fine-to-coarse ensemble. Fourth, MVP$_{obj}(seg\_tok, seg)$ achieve performances that are close to MVP$_{pair}(seg\_tok, char, e)$ on lcqmc, with less than half of the time in pre-training and fine-tuning. But the model's improvements on NER tasks are less significant. Notice that for pretraining and finetuning, MVP$_{pair}$ takes almost twice the resources and time than MVP$_{obj}$. Thus, for resources limited scenarios, MVP$_{obj}$ could be a more suitable choice. 

\subsection{Effects of relative importance coefficient}

\begin{figure}[h]
\begin{center}
%\framebox[4.0in]{$\;$}
\includegraphics[width=1.00\textwidth]{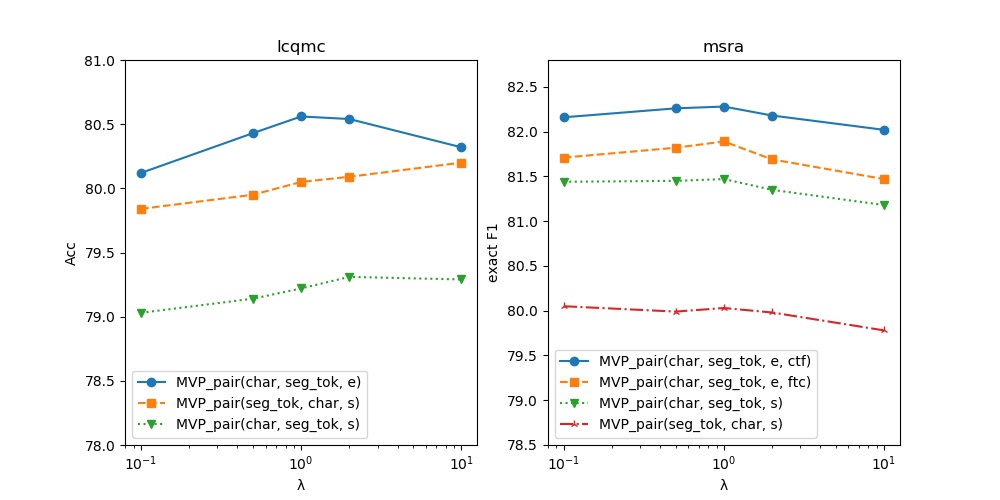}
\end{center}
\caption{How the value of coefficient $\lambda$ affects the models' downstream performances.}
\label{fig:ablation_on_lambda}
\end{figure}

In this section, we investigate how the relative importance coefficient $\lambda$ for MVP$_{pair}(char, seg\_tok)$ affects the ALBERT models' downstream performances. Here for MVP$_{pair}$, $\lambda$ is assigned to the loss term from coarse-grained vocab. We set $\lambda$ equal to 0.1, 0.5, 1, 2, 10, and run per-training under each of the setting. Finetuning results on \emph{lcqmc} and \emph{msra} are reported in Figure~\ref{fig:ablation_on_lambda}. For ensemble MVP$_{pair}(char, seg\_tok, e)$, $\lambda=1.0$ provides the best performances. And for MVP$_{pair}(char, seg\_tok, s)$, increasing $\lambda$ worsen the performance on NER, but not on sentence level tasks, and vice versa for MVP$_{pair}(seg\_tok, char, s)$.

\section{Conclusions}

In this work, we first propose a novel method, \emph{seg\_tok}, to re-build the vocabulary of Chinese BERT, with the help of Chinese word segmentation (CWS) and subword tokenization. Then we propose three versions of multi-vocabulary pretraining (MVP) to improve the models' performance.   Experiments show that: (a) compared with char based vocabulary, \emph{seg\_tok} does not only improves the performances of Chinese PLMs on sentence level tasks, it can also improve efficiency; (b) MVP improves PLMs' downstream performance, especially it can improve \emph{seg\_tok}'s performances on sequence labeling tasks.

%  select a few sentences

\bibliography{iclr2021_conference,emnlp2020}

\begin{thebibliography}{29}
\providecommand{\natexlab}[1]{#1}
\providecommand{\url}[1]{\texttt{#1}}
\expandafter\ifx\csname urlstyle\endcsname\relax
  \providecommand{\doi}[1]{doi: #1}\else
  \providecommand{\doi}{doi: \begingroup \urlstyle{rm}\Url}\fi

\bibitem[{Bowman} et~al.(2015){Bowman}, {Angeli}, {Potts}, and
  {Manning}]{bowman2015nli}
Samuel~R. {Bowman}, Gabor {Angeli}, Christopher {Potts}, and Christopher~D.
  {Manning}.
\newblock {A large annotated corpus for learning natural language inference}.
\newblock \emph{arXiv e-prints}, art. arXiv:1508.05326, August 2015.

\bibitem[{Clark} et~al.(2020){Clark}, {Luong}, {Le}, and {Manning}]{electra}
Kevin {Clark}, Minh-Thang {Luong}, Quoc~V. {Le}, and Christopher~D. {Manning}.
\newblock {ELECTRA: Pre-training Text Encoders as Discriminators Rather Than
  Generators}.
\newblock \emph{arXiv e-prints}, art. arXiv:2003.10555, March 2020.

\bibitem[Conneau et~al.(2018)Conneau, Rinott, Lample, Williams, Bowman,
  Schwenk, and Stoyanov]{conneau-etal-2018-xnli}
Alexis Conneau, Ruty Rinott, Guillaume Lample, Adina Williams, Samuel Bowman,
  Holger Schwenk, and Veselin Stoyanov.
\newblock {XNLI}: Evaluating cross-lingual sentence representations.
\newblock In \emph{Proceedings of the 2018 Conference on Empirical Methods in
  Natural Language Processing}, pp.\  2475--2485, Brussels, Belgium,
  October-November 2018. Association for Computational Linguistics.
\newblock \doi{10.18653/v1/D18-1269}.
\newblock URL \url{https://www.aclweb.org/anthology/D18-1269}.

\bibitem[{Cui} et~al.(2019{\natexlab{a}}){Cui}, {Che}, {Liu}, {Qin}, {Yang},
  {Wang}, and {Hu}]{chinese-bert-wwm}
Yiming {Cui}, Wanxiang {Che}, Ting {Liu}, Bing {Qin}, Ziqing {Yang}, Shijin
  {Wang}, and Guoping {Hu}.
\newblock {Pre-Training with Whole Word Masking for Chinese BERT}.
\newblock \emph{arXiv e-prints}, art. arXiv:1906.08101, June
  2019{\natexlab{a}}.

\bibitem[{Cui} et~al.(2019{\natexlab{b}}){Cui}, {Che}, {Liu}, {Qin}, {Yang},
  {Wang}, and {Hu}]{wwm19chinese}
Yiming {Cui}, Wanxiang {Che}, Ting {Liu}, Bing {Qin}, Ziqing {Yang}, Shijin
  {Wang}, and Guoping {Hu}.
\newblock {Pre-Training with Whole Word Masking for Chinese BERT}.
\newblock \emph{arXiv e-prints}, art. arXiv:1906.08101, June
  2019{\natexlab{b}}.

\bibitem[Devlin et~al.(2018)Devlin, Chang, Lee, and Toutanova]{devlin2018bert}
Jacob Devlin, Ming-Wei Chang, Kenton Lee, and Kristina Toutanova.
\newblock Bert: Pre-training of deep bidirectional transformers for language
  understanding.
\newblock \emph{arXiv preprint arXiv:1810.04805}, 2018.

\bibitem[Dong et~al.(2016)Dong, Zhang, Zong, Hattori, and
  Di]{lstm-radical-level-features}
C.~Dong, Jiajun Zhang, C.~Zong, M.~Hattori, and Hui Di.
\newblock Character-based lstm-crf with radical-level features for chinese
  named entity recognition.
\newblock In \emph{NLPCC/ICCPOL}, 2016.

\bibitem[Joshi et~al.(2019)Joshi, Chen, Liu, Weld, Zettlemoyer, and
  Levy]{joshi2019spanbert}
Mandar Joshi, Danqi Chen, Yinhan Liu, Daniel~S. Weld, Luke Zettlemoyer, and
  Omer Levy.
\newblock {SpanBERT}: Improving pre-training by representing and predicting
  spans.
\newblock \emph{arXiv preprint arXiv:1907.10529}, 2019.

\bibitem[Kingma \& Ba(2015)Kingma and Ba]{KingmaICML15}
Diederik~P. Kingma and Jimmy Ba.
\newblock Adam: A method for stochastic optimization.
\newblock In \emph{ICML}, 2015.

\bibitem[Lan et~al.(2019)Lan, Chen, Goodman, Gimpel, Sharma, and
  Soricut]{lan2019albert}
Zhenzhong Lan, Mingda Chen, Sebastian Goodman, Kevin Gimpel, Piyush Sharma, and
  Radu Soricut.
\newblock Albert: A lite bert for self-supervised learning of language
  representations.
\newblock \emph{arXiv preprint arXiv:1909.11942}, 2019.

\bibitem[Levow(2006)]{levow-2006-third}
Gina-Anne Levow.
\newblock The third international {C}hinese language processing bakeoff: Word
  segmentation and named entity recognition.
\newblock In \emph{Proceedings of the Fifth {SIGHAN} Workshop on {C}hinese
  Language Processing}, pp.\  108--117, Sydney, Australia, July 2006.
  Association for Computational Linguistics.
\newblock URL \url{https://www.aclweb.org/anthology/W06-0115}.

\bibitem[{Li} et~al.(2019){Li}, {Meng}, {Sun}, {Han}, {Yuan}, and
  {Li}]{is-word-seg-necessary}
Xiaoya {Li}, Yuxian {Meng}, Xiaofei {Sun}, Qinghong {Han}, Arianna {Yuan}, and
  Jiwei {Li}.
\newblock {Is Word Segmentation Necessary for Deep Learning of Chinese
  Representations?}
\newblock \emph{arXiv e-prints}, art. arXiv:1905.05526, May 2019.

\bibitem[{Liu} et~al.(2019{\natexlab{a}}){Liu}, {Zhou}, {Zhao}, {Wang}, {Ju},
  {Deng}, and {Wang}]{K-BERT}
Weijie {Liu}, Peng {Zhou}, Zhe {Zhao}, Zhiruo {Wang}, Qi~{Ju}, Haotang {Deng},
  and Ping {Wang}.
\newblock {K-BERT: Enabling Language Representation with Knowledge Graph}.
\newblock \emph{arXiv e-prints}, art. arXiv:1909.07606, September
  2019{\natexlab{a}}.

\bibitem[Liu et~al.(2018)Liu, Chen, Deng, Zeng, Chen, Li, and
  Tang]{liu-etal-2018-lcqmc}
Xin Liu, Qingcai Chen, Chong Deng, Huajun Zeng, Jing Chen, Dongfang Li, and
  Buzhou Tang.
\newblock {LCQMC}:a large-scale {C}hinese question matching corpus.
\newblock In \emph{Proceedings of the 27th International Conference on
  Computational Linguistics}, pp.\  1952--1962, Santa Fe, New Mexico, USA,
  August 2018. Association for Computational Linguistics.
\newblock URL \url{https://www.aclweb.org/anthology/C18-1166}.

\bibitem[{Liu} et~al.(2019{\natexlab{b}}){Liu}, {Ott}, {Goyal}, {Du}, {Joshi},
  {Chen}, {Levy}, {Lewis}, {Zettlemoyer}, and {Stoyanov}]{roberta}
Yinhan {Liu}, Myle {Ott}, Naman {Goyal}, Jingfei {Du}, Mandar {Joshi}, Danqi
  {Chen}, Omer {Levy}, Mike {Lewis}, Luke {Zettlemoyer}, and Veselin
  {Stoyanov}.
\newblock {RoBERTa: A Robustly Optimized BERT Pretraining Approach}.
\newblock \emph{arXiv e-prints}, art. arXiv:1907.11692, July
  2019{\natexlab{b}}.

\bibitem[Peters et~al.(2018)Peters, Neumann, Iyyer, Gardner, Clark, Lee, and
  Zettlemoyer]{Peters2018elmo}
Matthew~E. Peters, Mark Neumann, Mohit Iyyer, Matt Gardner, Christopher Clark,
  Kenton Lee, and Luke Zettlemoyer.
\newblock Deep contextualized word representations.
\newblock In \emph{Proc. of NAACL}, 2018.

\bibitem[Radford et~al.(2018)Radford, Narasimhan, Salimans, and
  Sutskever]{radford2018language}
Alec Radford, Karthik Narasimhan, Tim Salimans, and Ilya Sutskever.
\newblock Improving language understanding by generative pre-training.
\newblock \emph{arXiv e-prints}, 2018.

\bibitem[Radford et~al.(2019)Radford, Wu, Child, Luan, Amodei, and
  Sutskever]{radford2019language}
Alec Radford, Jeff Wu, Rewon Child, David Luan, Dario Amodei, and Ilya
  Sutskever.
\newblock Language models are unsupervised multitask learners.
\newblock \emph{arXiv e-prints}, 2019.

\bibitem[{Rajpurkar} et~al.(2018){Rajpurkar}, {Jia}, and {Liang}]{squad}
Pranav {Rajpurkar}, Robin {Jia}, and Percy {Liang}.
\newblock {Know What You Don't Know: Unanswerable Questions for SQuAD}.
\newblock \emph{arXiv e-prints}, art. arXiv:1806.03822, June 2018.

\bibitem[Sun et~al.(2019)Sun, Wang, Li, Feng, Chen, Zhang, Tian, Zhu, Tian, and
  Wu]{sun2019ernie}
Yu~Sun, Shuohuan Wang, Yukun Li, Shikun Feng, Xuyi Chen, Han Zhang, Xin Tian,
  Danxiang Zhu, Hao Tian, and Hua Wu.
\newblock Ernie: Enhanced representation through knowledge integration.
\newblock \emph{arXiv preprint arXiv:1904.09223}, 2019.

\bibitem[{Sun} et~al.(2019){Sun}, {Wang}, {Li}, {Feng}, {Tian}, {Wu}, and
  {Wang}]{ERNIE2}
Yu~{Sun}, Shuohuan {Wang}, Yukun {Li}, Shikun {Feng}, Hao {Tian}, Hua {Wu}, and
  Haifeng {Wang}.
\newblock {ERNIE 2.0: A Continual Pre-training Framework for Language
  Understanding}.
\newblock \emph{arXiv e-prints}, art. arXiv:1907.12412, July 2019.

\bibitem[Vaswani et~al.(2017)Vaswani, Shazeer, Parmar, Uszkoreit, Jones, Gomez,
  Kaiser, and Polosukhin]{vaswani2017attention}
Ashish Vaswani, Noam Shazeer, Niki Parmar, Jakob Uszkoreit, Llion Jones,
  Aidan~N Gomez, {\L}ukasz Kaiser, and Illia Polosukhin.
\newblock Attention is all you need.
\newblock In \emph{NIPS}, 2017.

\bibitem[{Wang} et~al.(2019){Wang}, {Bi}, {Yan}, {Wu}, {Bao}, {Xia}, {Peng},
  and {Si}]{StructBERT}
Wei {Wang}, Bin {Bi}, Ming {Yan}, Chen {Wu}, Zuyi {Bao}, Jiangnan {Xia}, Liwei
  {Peng}, and Luo {Si}.
\newblock {StructBERT: Incorporating Language Structures into Pre-training for
  Deep Language Understanding}.
\newblock \emph{arXiv e-prints}, art. arXiv:1908.04577, August 2019.

\bibitem[Xu et~al.(2015)Xu, Chen, Xia, Lu, Liu, and
  Wang]{Word-Embedding-Composition}
Ruifeng Xu, Tao Chen, Yunqing Xia, Qin Lu, Bin Liu, and Xuan Wang.
\newblock Word embedding composition for data imbalances in sentiment and
  emotion classification.
\newblock \emph{Cognitive Computation}, 7, 02 2015.
\newblock \doi{10.1007/s12559-015-9319-y}.

\bibitem[{Yang} et~al.(2019){Yang}, {Dai}, {Yang}, {Carbonell},
  {Salakhutdinov}, and {Le}]{xlnet}
Zhilin {Yang}, Zihang {Dai}, Yiming {Yang}, Jaime {Carbonell}, Ruslan
  {Salakhutdinov}, and Quoc~V. {Le}.
\newblock {XLNet: Generalized Autoregressive Pretraining for Language
  Understanding}.
\newblock \emph{arXiv e-prints}, art. arXiv:1906.08237, June 2019.

\bibitem[Yin et~al.(2016)Yin, Wang, Li, Li, and Wang]{yin-etal-2016-multi}
Rongchao Yin, Quan Wang, Peng Li, Rui Li, and Bin Wang.
\newblock Multi-granularity {C}hinese word embedding.
\newblock In \emph{Proceedings of the 2016 Conference on Empirical Methods in
  Natural Language Processing}, pp.\  981--986, Austin, Texas, November 2016.
  Association for Computational Linguistics.
\newblock \doi{10.18653/v1/D16-1100}.
\newblock URL \url{https://www.aclweb.org/anthology/D16-1100}.

\bibitem[{Zaheer} et~al.(2020){Zaheer}, {Guruganesh}, {Dubey}, {Ainslie},
  {Alberti}, {Ontanon}, {Pham}, {Ravula}, {Wang}, {Yang}, and {Ahmed}]{bigbird}
Manzil {Zaheer}, Guru {Guruganesh}, Avinava {Dubey}, Joshua {Ainslie}, Chris
  {Alberti}, Santiago {Ontanon}, Philip {Pham}, Anirudh {Ravula}, Qifan {Wang},
  Li~{Yang}, and Amr {Ahmed}.
\newblock {Big Bird: Transformers for Longer Sequences}.
\newblock \emph{arXiv e-prints}, art. arXiv:2007.14062, July 2020.

\bibitem[Zhao et~al.(2013)Zhao, Utiyama, Sumita, and Lu]{zhao2013empirical}
Hai Zhao, Masao Utiyama, Eiichiro Sumita, and Bao-Liang Lu.
\newblock An empirical study on word segmentation for chinese machine
  translation.
\newblock In \emph{International Conference on Intelligent Text Processing and
  Computational Linguistics}, pp.\  248--263. Springer, 2013.

\bibitem[Zou et~al.(2013)Zou, Socher, Cer, and
  Manning]{zou-etal-2013-bilingual}
Will~Y. Zou, Richard Socher, Daniel Cer, and Christopher~D. Manning.
\newblock Bilingual word embeddings for phrase-based machine translation.
\newblock In \emph{Proceedings of the 2013 Conference on Empirical Methods in
  Natural Language Processing}, pp.\  1393--1398, Seattle, Washington, USA,
  October 2013. Association for Computational Linguistics.
\newblock URL \url{https://www.aclweb.org/anthology/D13-1141}.

\end{thebibliography}
\bibliographystyle{iclr2021_conference}

\end{document}